# AIRCRAFT FUSELAGE DEFECT DETECTION USING DEEP NEURAL NETWORKS

*Touba Malekzadeh, Milad Abdollahzadeh, Hossein Nejati, Ngai-Man Cheung*

Singapore University of Technology and Design (SUTD)

## ABSTRACT

To ensure flight safety of aircraft structures, it is necessary to have regular maintenance using visual and nondestructive inspection (NDI) methods. In this paper, we propose an automatic image-based aircraft defect detection using Deep Neural Networks (DNNs). To the best of our knowledge, this is the first work for aircraft defect detection using DNNs. We perform a comprehensive evaluation of state-of-the-art feature descriptors and show that the best performance is achieved by vgg-f DNN as feature extractor with a linear SVM classifier. To reduce the processing time, we propose to apply SURF key point detector to identify defect patch candidates. Our experiment results suggest that we can achieve over 96% accuracy at around 15s processing time for a high-resolution (20-megapixel) image on a laptop.

## 1. INTRODUCTION

Aircraft inspection and maintenance is an essential to safe air transportation [1-3]. A fully automated system to monitor the structural health of an aircraft has the potential to reduce operating costs, increase flight safety and improve aircraft availability [4]. This paper makes a contribution to the field of automatic defect detection of an aircraft fuselage with computer vision techniques.

In [5], they research computer-simulated visual inspection (VI) and non-destructive inspection (NDI) tasks. However, these visual inspection tasks were performed by human inspectors who searched for defect manually. Our proposed algorithm is a completely automatic inspection.

In recent years, deep neural networks (DNN) have shown promising results in different classification tasks [6-10]. Although DNNs can be used to perform classification directly using the output of the last network layer, they can also be used as a feature extractor combined with a classifier [11].

In this paper, we investigate a classification system that employs a DNN, pretrained using natural images, to extract features suitable to another domain, i.e., aircraft fuselage defect detection, where there are few samples available. The contributions of this study are:

1) To the best of our knowledge, this is the first work for automatic defect detection of aircraft fuselage using visual images and deep learning.
2) We propose a fast and accurate detection algorithm with selection of region of interest using SURF interest point extractor.
3) We propose techniques to handle washed and unwashed fuselage based on pre- and post-processing.

To the best of our knowledge, there is no previous work on automatic image-based aircraft defect detection. Image based defect detection has been investigated for other problems: In [12] X-ray images of metallic components are used as a non-destructive testing method, to detect the defects within casting components. In [13] they propose a deep convolutional neural network solution to the analysis of image data for the detection of rail surface defects. The images were obtained from many hours of automated video recordings. However, the image and defect characteristics of these problems are rather different from ours.

The rest of the paper is organized as follows. In Section 2 we give a detailed explanation of our datasets. Section 3 explains the proposed algorithm and the DNN-derived features generated automatically from a dataset of fuselage images. The performance evaluation of the proposed algorithm is provided in Section 4, while Section 5 presents the conclusion and the feature work.

## 2. DATASETS

Our dataset images are taken in a straight view of the airplane fuselage. During the inspection, a drone can be used to capture these images automatically. Images are stored in JPEG format. All images have three color channels and 3888×5184 resolution. Some examples of aircraft fuselage images with defects are illustrated in Figure 1. For each image, a binary mask is created by an experienced inspector to represent defects. Considering $S_D$ as the set of defect pixels in an image, corresponding mask M is created as follows:

$$M(i,j) = \begin{cases} 1 & I(i,j) \in S_D \\ 0 & otherwise \end{cases} \quad (1)$$

In (1), $M(i,j)$ and $I(i,j)$ represents pixel value in $i^{th}$ row and $j^{th}$ column of mask and input image respectively.

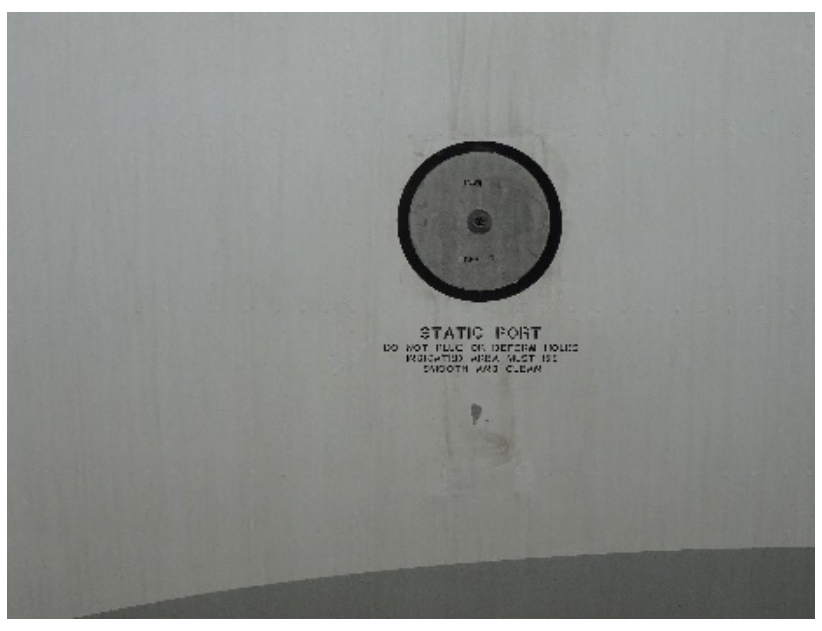

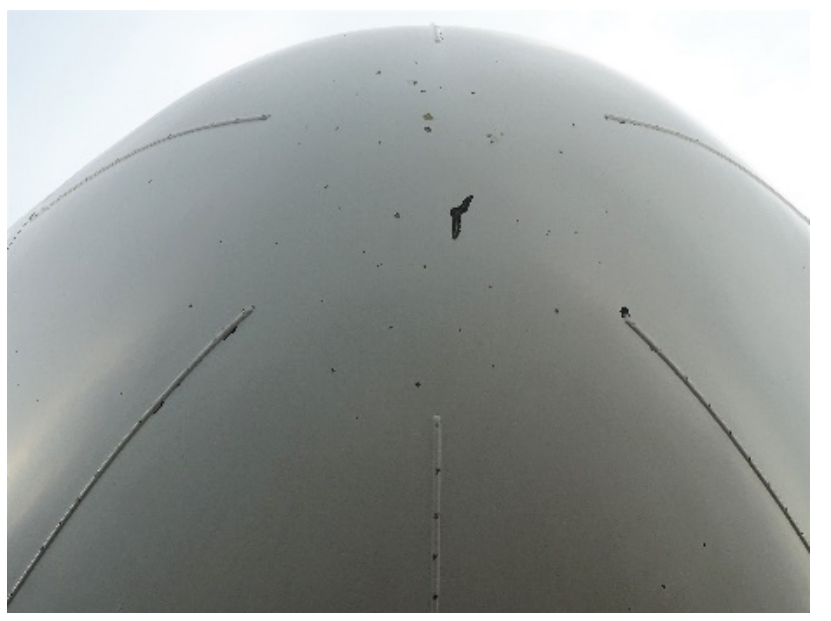
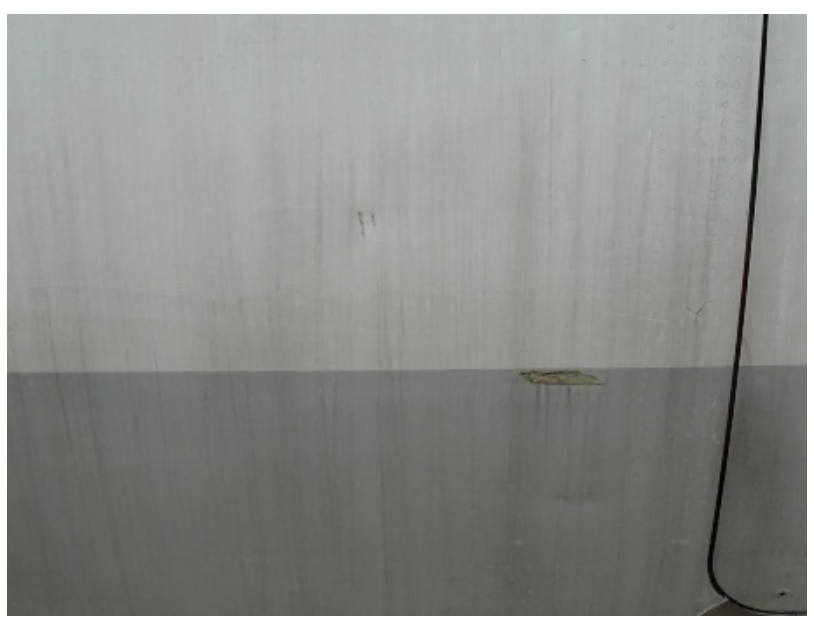

Figure 1. Examples of defects in real images of fuselage

Each image is partitioned into the 65×65 patches with respect to image resolution to include the smallest defect within a single patch. There are two classes of patches: defect and no-defect. A patch is labeled as defect if the majority of its pixels belong to the $S_D$, otherwise it is labeled as a no-defect patch. Figure 2 shows some examples of the defect and no-defect patches in our dataset. There are two challenges for this application. First, some defect patches and no-defect patches have similar appearance. As one can see there are some defect patches that can be easily detected (i.e. sharp intensity change in a uniform background). However, there are some homogeneous defect patches that are very similar to no-defect patterns. On the other hand, there are some no-defect parts of the fuselage (e.g. riveted joints) that are similar to defects. Second, our dataset is unbalance as the important class (defect class) is the minority one. Hence, a method is required to improve the minority class recognition rate. We apply a data-level balancing method [14] to improve the training set via both oversampling the minority class (defect patches) and undersampling the majority class (no-defect patches). Oversampling is done by augmenting the minority class and undersampling is done by randomly removal of samples in the majority class. Data augmentation is done by rotating the original defect patches at 6 different angles ($0^\circ$, $60^\circ, 120^\circ, \ldots, 300^\circ$) and also flipping vertically and horizontally.

## 3. METHODOLOGY

In this work, we propose a patch-based scheme for detection of defects. Specifically, we partition the image into 65x65 patches and classify each patch into defect / non-defect class. The classification is a two-step process: First, we compute a set of features for each patch; second, we build a classification model based on the extracted features.

For computing the discriminative features, we evaluate and compare a set of techniques including deep neural network, local descriptors, and texture features. Our experiments show that using the pretrained convolutional neural network (CNN) results in the best performance. For the classification step, we use SVM with a linear kernel.

In the experiment, we split the data into disjoint training and testing sets, in a manner that the data which is present in the training set is not allowed to be in the testing set. But in order to make these two sets completely disjoint, we employ 10-fold cross validation on the images rather than the patches, i.e. the patches of a particular image have the same cross-validation index as their parent image. This approach prevents having highly correlated data in both training and testing sets, that results in high accuracy which is not the case.

Considering the high resolution of images in our dataset and the maximum allowed size of a patch, it needs high computational complexity to evaluate all the patches in a single image. Since this work addresses an industrial application, processing time is a critical factor. We propose to boost our algorithm by analyzing the region of interest which is extracted by SURF [15] detector.

Another critical problem in defect detection of airplane fuselage is about washed or unwashed fuselage. Unwashed fuselage with dirt causes some problems in detection of defects. We propose to extend our algorithm for both washed and unwashed conditions.

### 3.1. Feature Extraction

As we will show in the results, using CNN as feature extractor achieves the best performance. Therefore, we have used a convolutional neural network (CNN) pre-trained on ImageNet as a feature extractor for our dataset. Transferring the knowledge of an existing CNN to a new domain has been studied and proved successful in several applications [11, 16, 17]. This approach is more appropriate for our application rather than fine-tuning the CNN [18] due to several reasons such as size and types of our dataset. Considering the limited size of our dataset, we propose to build a classifier model on top of the output (activations) of the hidden layers [19]. Furthermore, since the dataset (ImageNet) that was used to train CNN is quite different from our dataset, it is better to use the activations of the earlier layers of the network to construct the classifier. The block diagram of the proposed method for defect detection is shown in figure 3. As discussed in section 2, an equally balanced set of patches is used for training.

In this work, we evaluate two CNN models related to ImageNet: AlexNet [6] and VGG-F networks [7]. As illustrated in figure 3, the nets comprise of eight layers; the first five are convolutional layers and the remaining three are fully connected layers. The size of the descriptors is 4096 for ‘fc6’ and ‘fc7’ layers and 1000 for ‘fc8’ layer. The input image of these models are images of 244 ×244 ×3 pixels. For this reason, our 65 × 65 pixel-patches were resized to the required size (all three channels are equal). Considering K neurons in fully connected layers, we consider the extracted layer as a feature vector $\boldsymbol{f} = [f_1,\ f_2, \ldots, f_K]$. After obtaining deep feature vectors for each patch, SVM with linear kernel is used for classification. Based on our experimental results we adapt fc6 of VGG-F as feature extractor.

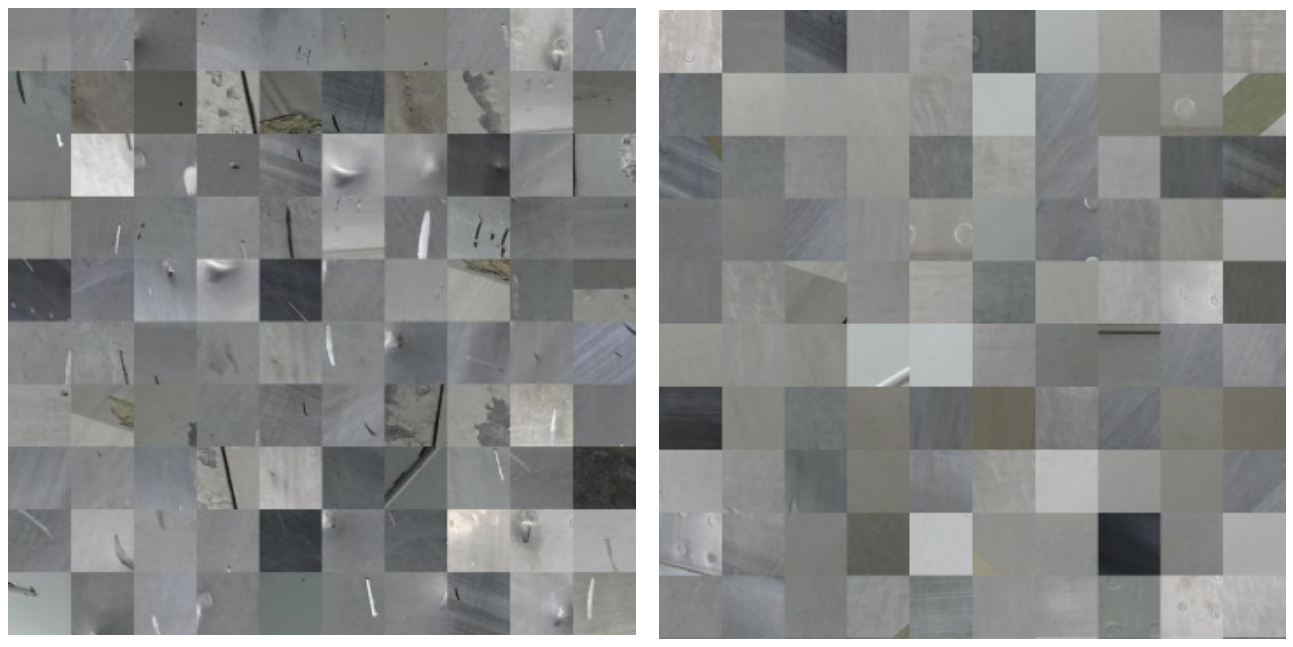

Figure 2. Examples of defect (left) and no-defect patches (right)

### 3.2. Boosting Defect Detection

As discussed in 3.1, the SVM classifier is fed with a set of discriminative features, extracted for each patch. Considering our high-resolution images (20-megapixel), there are lots of patches to be evaluated by the feature extractor and the classifier. To speed up the processing, one approach is to decrease the number of patches by increasing the patch size, but this affects the accuracy of detection. In general, increasing the patch size reduces the computational complexity but degrades the accuracy and vice versa. Therefore, there is a trade-off between the accuracy and the computational complexity to choose patch size. We have tested a variation of patch sizes from 20x20 to 100x100 pixels and the best results are achieved by patch size 65x65 pixels.

Considering this patch size and the resolution of our images, it is a time-consuming task to evaluate all the patches within an image. We propose to boost our algorithm via enforcing the evaluation to some regions of interest. The regions of interest must include all the probable defect areas.

Through our experiments, we found that, in most images, speeded up robust feature (SURF) is able to detect all the defect regions together with some normal regions which are similar to the defects. Therefore, we propose to apply SURF interest point detector to select some patches to be included in the evaluation procedure. A patch is included in the defect evaluation procedure if it contains at least one SURF key point. In this way, lots of homogenous regions of the fuselage are excluded from the evaluation step. Our results show that evaluating only the selected patches of the regions of interest can boost defect detection algorithm by a 6x speed-up. Figure 4 shows the block diagram of the boosted defect detection algorithm.

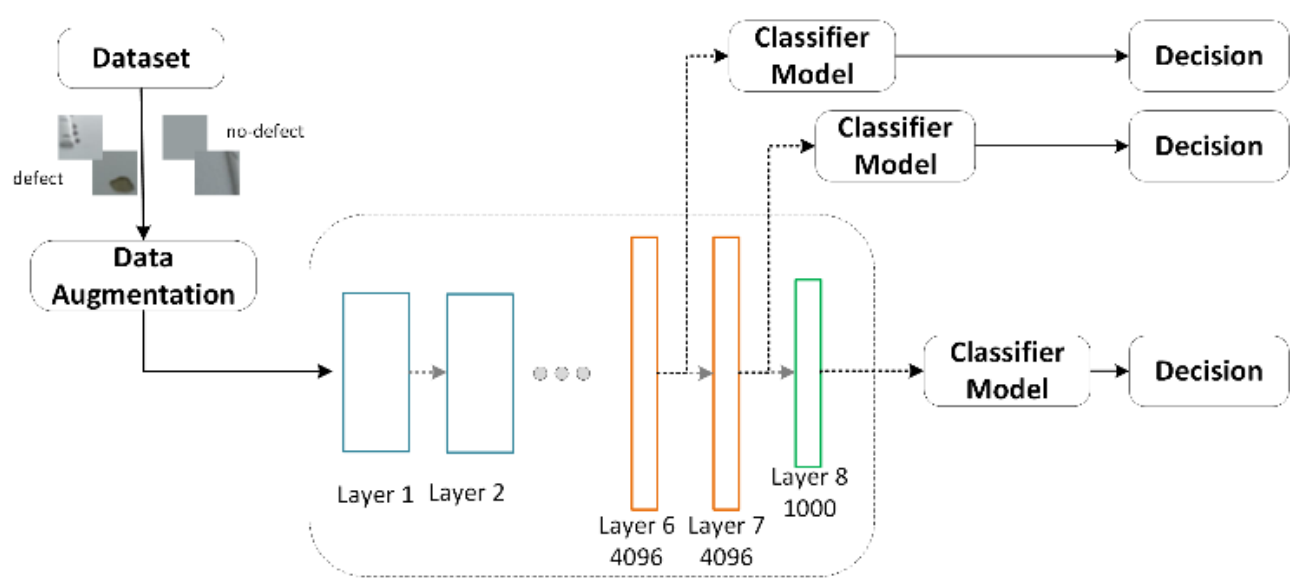

Figure 3. Block diagram of the proposed method

### 3.3. Post-processing

As this work is an industrial application, we have to make the algorithm applicable for different conditions. Washing status of the aircraft is an important factor, which affects the defect detection procedure. As aircraft exterior cleaning procedure is time and effort consuming, it is usually done occasionally. As a result, an aircraft could be unwashed with dirty spots on it which mislead the defect detection process. In order to overcome this issue, we employ a user interface to choose between two different conditions of washed or unwashed aircraft. In the condition of washed aircraft, the detection pipeline is the same as discussed in section 3.2. But for an unwashed aircraft with dirty spots on the fuselage, we propose to apply a low-pass Gaussian filter to reduce the noise-like spots on the fuselage images, in a manner that has the minimum smoothing effect on the real defects. But, there is a trade-off between reducing the high-pass components of the image and retaining the defects thoroughly. Our approach to overcome this problem is to have a post-processing after classification of the patches, which is done by a similarity comparison of the adjacent patches. The intensity range of the patch is used as the similarity metric. Specifically, if a patch is detected as a defect patch, it is most likely to be in a defect region, so we also test its adjacent patches with a post-processing scheme to ensure all the patches in a defect region are detected. If an adjacent patch satisfies the similarity threshold, it is classified into the defect class.

To measure the similarity, we assess the adjacent patches with a low computational complexity metric. Intensity variation is used as the similarity metric as follows:

$$IV(i,j) = \left|\left(I_{max}^{i} - I_{min}^{i}\right) - \left(I_{max}^{j} - I_{min}^{j}\right)\right| \quad (2)$$

Where $I_{max}^{i}$ and $I_{min}^{i}$ denote maximum and minimum values of intensity in $i^{th}$ patch. Adjacent defect patches have shown to be strongly correlated based on this metric. Our choice of intensity variation as similarity metric here is merely meant to be as a proof of concept, and there may be better similarity metrics that may lead to better performance.

The intuition of this scheme is that the defect patches cluster in almost all the cases and in the worst situation, and our model is able to detect at least one defect patch of the defect cluster. This post-processing scheme makes the results comparable with the results of the main pipeline and improves the sensitivity of the system significantly.

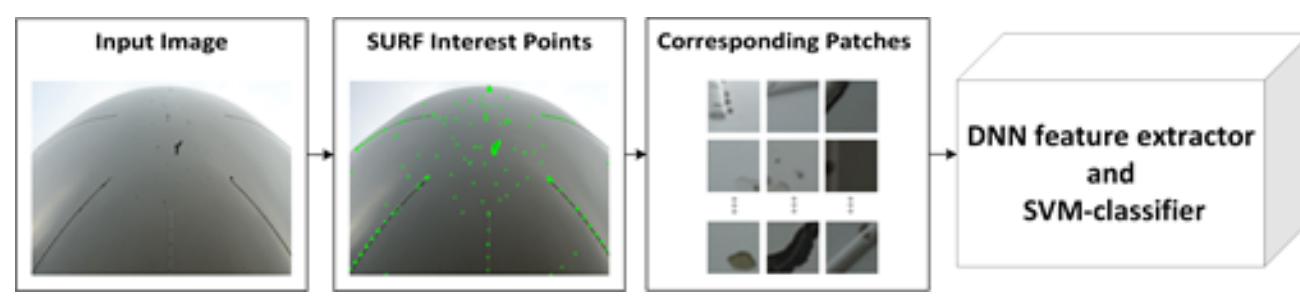

Figure 4. Block diagram of the proposed fast detection method

## 4. RESULTS

We have implemented our algorithm on MATLAB. The training is done on a machine with Intel Xeon CPU and 64 GB RAM. The testing running time is measured on a laptop with Intel Core i7 CPU with 2.40 GHz and 8 GB RAM. In order to extract features of the pretrained CNNs, we have used available Matconvnet library [20].

### 4.1. Comparing Features

In order to choose the appropriate feature extractor, we have compared a set of well-known feature descriptors in the literature including RGB histogram, HSV histogram, local binary pattern [21] (LBP), SURF and VGG-f and AlexNet pretrained CNNs. In each case, feature descriptor is applied to the patches and results are classified with linear SVM. For pretrained networks, the classification is performed on the activations on fully connected layers. Table 1 reports the average results of applying these feature descriptors on our data set. Results are evaluated in terms of sensitivity (i.e., Sens = TP/(TP+FN)), specificity (i.e., Spec = TN/(TN+FP)), and accuracy (i.e., Acc= (TP+TN) / (TP+TN +FP+FN)).

Table.1 Average performance of various feature descriptors across all folds.

| Method | | Accuracy | Sensitivity | Specificity |
|---|---|---|---|---|
| RGB histogram | | 0.603722 | 0.295050 | 0.808990 |
| HSV histogram | | 0.602995 | 0.309751 | 0.798006 |
| LBP | | 0.603833 | 0.126360 | 0.921346 |
| SURF | | 0.636679 | 0.274245 | 0.846598 |
| VGG-f | **$FC^6$** | **0.876236** | **0.854368** | 0.905322 |
| | $FC^7$ | 0.875025 | 0.849498 | 0.908975 |
| | $FC^8$ | 0.871628 | 0.848207 | 0.902778 |
| AlexNet | $FC^6$ | 0.847333 | 0.711691 | 0.937537 |
| | $FC^7$ | 0.846318 | 0.706273 | **0.939451** |
| | $FC^8$ | 0.834154 | 0.683291 | 0.934481 |

The generalization error of the classifier models is estimated using the 10-fold cross-validation. In each trial of 10-fold cross-validation, 9 folds are used for training and the last fold is used for evaluation. This process is repeated 10 times, leaving one different fold for evaluation each time. As one can see, using pre-trained CNNs as feature extractor results in better results. Based on the results, we have chosen the sixth fully connected layer of VGG-f as our feature extractor.

### 4.2. Results of Testing on Unseen Images

In this section, we have provided the results of applying our boosted defect detection algorithm (with SURF key-point detector) on a set of twelve unseen aircraft images. As described earlier this approach selects candidates for classification based on the SURF key-point detection results. Based on the results of 4.1, we have used fc6 of VGG-f as feature extractor. Table 2 shows the average results of applying the proposed algorithm on 12 unseen test images.

Table.2 Average performance of the proposed algorithm on unseen images

| Accuracy | Sensitivity | Specificity | Runtime (sec) |
|---|---|---|---|
| 0.963784 | 0.964891 | 0.963823 | 15.7874 |

Considering the results presented in Table 2, the proposed algorithm achieves about 96.37% accuracy, which means that only 3.63 % of the patches are misclassified. And also sensitivity is about 96.48% which indicates that only 3.52% of the defect patches are missed. It is worth noting that by analyzing the results visually, we found that every defect region is at least partially detected (although some patches in the region are missed in some rare situations). Therefore, practically the system has located all defect regions. As specificity of about 96 % means that approximately only 4 % of the whole airplane structure needs to be manually inspected by the worker, that is a significant saving of the worker effort and time. Also, the average run time is 15.78 seconds for an image with 3888×5184 resolution. This enables efficient automatic inspection.

## 5. CONCLUSION

In this paper, we proposed an automatic aircraft fuselage defect detection method based on DNNs and transfer learning. We evaluated multiple feature extractions and found that transferring features from pre-trained CNNs is a solution for defect detection. We proposed to boost defect detection algorithm using region of interest detected by SURF, and this has achieved 6x speedup comparing with a baseline approach that tests all patches. Proposed algorithm is able to detect almost all the defects of the aircraft fuselage and reduce the workload of manual inspection significantly.
In future, we plan to research other DNN architecture for this application.

## 6. REFERENCES

[1] C. G. Drury, P. Prabhu, and A. Gramopadhye, "Task analysis of aircraft inspection activities: methods and findings," in *Proceedings of the Human Factors Society Annual Meeting*, 1990, pp. 1181-1185.

[2] D. E. Bray and R. K. Stanley, *Nondestructive evaluation: a tool in design, manufacturing and service*: CRC press, 1996.

[3] K. A. Latorella and P. V. Prabhu, "A review of human error in aviation maintenance and inspection," *International Journal of Industrial Ergonomics,* vol. 26, pp. 133-161, 2000.

[4] R. Dalton, P. Cawley, and M. Lowe, "The potential of guided waves for monitoring large areas of metallic aircraft fuselage structure," *Journal of Nondestructive Evaluation,* vol. 20, pp. 29-46, 2001.

[5] K. Latorella, A. Gramopadhye, P. Prabhu, C. Drury, M. Smith, and D. Shanahan, "Computer-simulated aircraft inspection tasks for off-line experimentation," in *Proceedings of the Human Factors and Ergonomics Society Annual Meeting*, 1992, pp. 92-96.

[6] A. Krizhevsky, I. Sutskever, and G. E. Hinton, "Imagenet classification with deep convolutional neural networks," in *Advances in neural information processing systems*, 2012, pp. 1097-1105.

[7] K. Chatfield, K. Simonyan, A. Vedaldi, and A. Zisserman, "Return of the devil in the details: Delving deep into convolutional nets," *arXiv preprint arXiv:1405.3531,* 2014.

[8] Y. Zhou, S. Song, and N.-M. Cheung, "On Classification of Distorted Images with Deep Convolutional Neural Networks," *arXiv preprint arXiv:1701.01924,* 2017.

[9] D. Ciregan, U. Meier, and J. Schmidhuber, "Multi-column deep neural networks for image classification," in *Computer Vision and Pattern Recognition (CVPR), 2012 IEEE Conference on*, 2012, pp. 3642-3649.

[10] Y. Guo, H. Nejati, and N.-M. Cheung, "Deep neural networks on graph signals for brain imaging analysis," *arXiv preprint arXiv:1705.04828,* 2017.

[11] A. Sharif Razavian, H. Azizpour, J. Sullivan, and S. Carlsson, "CNN features off-the-shelf: an astounding baseline for recognition," in *Proceedings of the IEEE Conference on Computer Vision and Pattern Recognition Workshops*, 2014, pp. 806-813.

[12] D. Mery and C. Arteta, "Automatic Defect Recognition in X-Ray Testing Using Computer Vision," in *Applications of Computer Vision (WACV), 2017 IEEE Winter Conference on*, 2017, pp. 1026-1035.

[13] S. Faghih-Roohi, S. Hajizadeh, A. Núñez, R. Babuska, and B. De Schutter, "Deep convolutional neural networks for detection of rail surface defects," in *Neural Networks (IJCNN), 2016 International Joint Conference on*, 2016, pp. 2584-2589.

[14] B. Krawczyk, "Learning from imbalanced data: open challenges and future directions," *Progress in Artificial Intelligence,* vol. 5, pp. 221-232, 2016.

[15] H. Bay, T. Tuytelaars, and L. Van Gool, "Surf: Speeded up robust features," *Computer vision–ECCV 2006,* pp. 404-417, 2006.

[16] Y. Zhou, H. Nejati, T.-T. Do, N.-M. Cheung, and L. Cheah, "Image-based Vehicle Analysis using Deep Neural Network: A Systematic Study," *arXiv preprint arXiv:1601.01145,* 2016.

[17] H. Nejati, V. Pomponiu, T.-T. Do, Y. Zhou, S. Iravani, and N.-M. Cheung, "Smartphone and Mobile Image Processing for Assisted Living: Health-monitoring apps powered by advanced mobile imaging algorithms," *IEEE Signal Processing Magazine,* vol. 33, pp. 30-48, 2016.

[18] J. Yosinski, J. Clune, Y. Bengio, and H. Lipson, "How transferable are features in deep neural networks?," in *Advances in neural information processing systems*, 2014, pp. 3320-3328.

[19] V. Pomponiu, H. Nejati, and N.-M. Cheung, "Deepmole: Deep neural networks for skin mole lesion classification," in *Image Processing (ICIP), 2016 IEEE International Conference on*, 2016, pp. 2623-2627.

[20] A. Vedaldi and K. Lenc, "Matconvnet: Convolutional neural networks for matlab," in *Proceedings of the 23rd ACM international conference on Multimedia*, 2015, pp. 689-692.

[21] Z. Guo, L. Zhang, and D. Zhang, "A completed modeling of local binary pattern operator for texture classification," *IEEE Transactions on Image Processing,* vol. 19, pp. 1657-1663, 2010.